\documentclass[10pt,twocolumn,letterpaper]{article}

\usepackage[pagenumbers]{wacv} 

\usepackage{graphicx}
\usepackage{amsmath}
\usepackage{amssymb}
\usepackage{booktabs}

\usepackage{algorithm}
\usepackage{algpseudocode}
\usepackage{multirow}
\usepackage{diagbox}

\usepackage[pagebackref,breaklinks,colorlinks]{hyperref}

\usepackage[capitalize]{cleveref}
\crefname{section}{Sec.}{Secs.}
\Crefname{section}{Section}{Sections}
\Crefname{table}{Table}{Tables}
\crefname{table}{Tab.}{Tabs.}

\def\wacvPaperID{2076} 
\def\confName{WACV}
\def\confYear{2025}

\begin{document}

\title{LiLMaps: Learnable Implicit Language Maps}

\author{Evgenii Kruzhkov\\
Autonomous Intelligent Systems, Computer Science
Institute VI\\
University of Bonn, Germany\\
{\tt\small ekruzhkov@ais.uni-bonn.de}
\and
Sven Behnke\\
Autonomous Intelligent Systems, Computer Science
Institute VI -- Intelligent Systems and Robotics\\Center for Robotics
and the Lamarr Institute for Machine Learning and Artificial Intelligence\\
University of Bonn, Germany\\
{\tt\small behnke@cs.uni-bonn.de}
}
\maketitle

\begin{abstract}
One of the current trends in robotics is to employ large language models (LLMs) to provide non-predefined command execution and natural human-robot interaction.
It is useful to have an environment map together with its language representation, which can be further utilized by LLMs.
Such a comprehensive scene representation enables numerous ways of interaction with the map for autonomously operating robots.
In this work, we present an approach that enhances incremental implicit mapping through the integration of vision-language features.
Specifically, we (i) propose a decoder optimization technique for implicit language maps which can be used when new objects appear on the scene, and (ii) address the problem of inconsistent vision-language predictions between different viewing positions.
Our experiments demonstrate the effectiveness of LiLMaps and solid improvements in performance.
\end{abstract}


\section{Introduction}\label{sec:intro}
Classic robotic maps are commonly used for estimating distances to obstacles and costs of motions in navigation and localization tasks.
However, more comprehensive tasks, as well as natural human-robot interaction, may require a deeper understanding of the environment, and thus imply more advanced map representations.
For example, visual-language navigation is the task where a robot must interpret a natural language command from a non-expert user and proceed towards the goal according to the command. 
The environment might be unknown in advance, but the robot still must navigate in the shortest possible time.
From this example, it is clear that a map that allows one to easily find correlation between the given language command and a partially or fully mapped environment can be more useful than a pure obstacle costmap.

\begin{figure}[t]
  \centering
  \includegraphics[width=1.\linewidth]{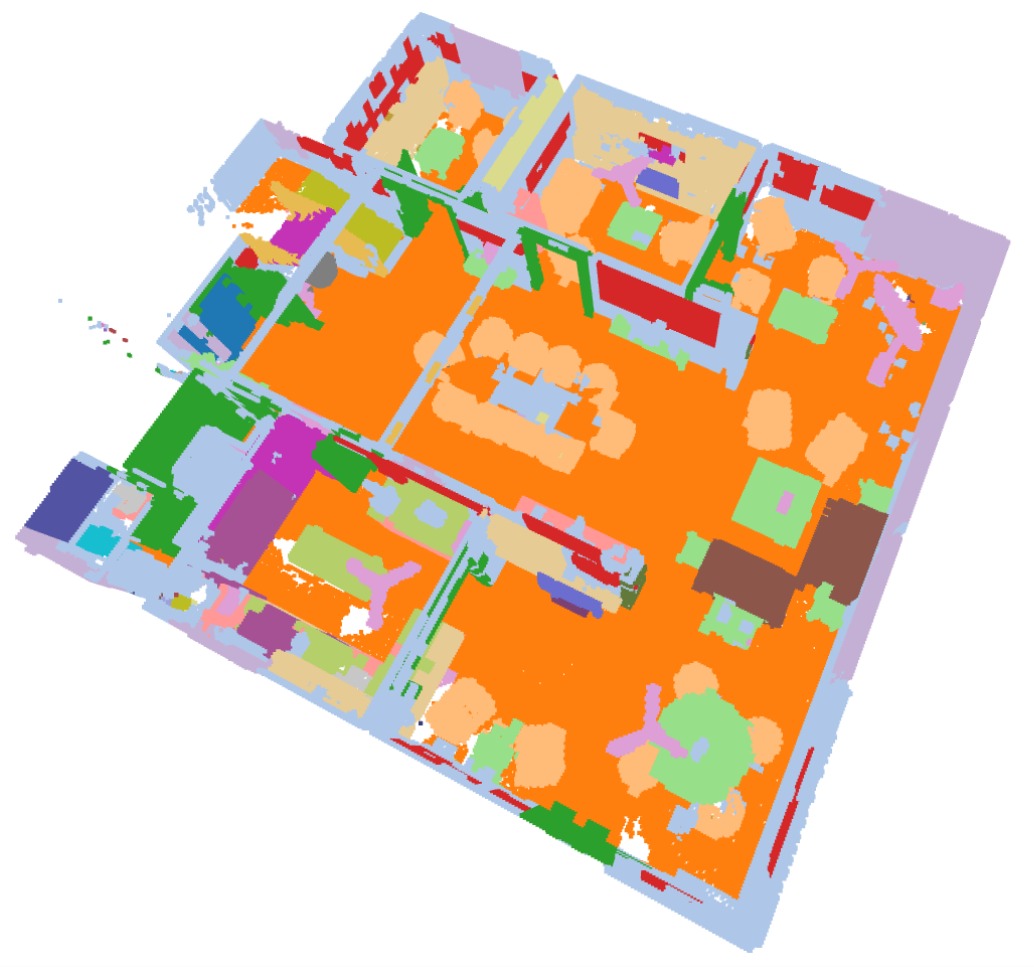}

   \caption{Reconstructed implicit language map built with LiLMaps. Semantic colors are assigned based on the similarity of reconstructed language features and CLIP\cite{radford2021learning} encodings of semantic categories from the Matterport3D dataset\cite{Matterport3D}.}
   \label{fig:intro}
\end{figure}

In this work, we make a step towards creation of efficient yet compact natural language environment representations and introduce Learnable implicit Language Maps (LiLMaps).
We chose an implicit representation because of its ability to compactly represent the data and for the possibility of further detailed reconstruction.

Recent studies in implicit mapping demonstrate outstanding results in geometry reconstruction.
While in these studies, geometry decoders can be easily pre-trained or even trained in the first few iterations, in our work we demonstrate that LiLMaps performs better compared to the pre-trained language decoders because some language features can be poorly represented in them.
Moreover, pre-training the decoder to represent every possible language feature could make its structure and training process significantly more complicated, and it can reduce its flexibility for different applications.

Another challenging problem that frequently appears in incremental language learning is the inconsistency of measurements taken from different viewing positions.
Precise range sensors, such as LiDARs and RGB-D cameras, are commonly used in implicit mapping, but usually they do not provide contradictory measurements.
However, in visual-language navigation tasks, information about the environment is often derived from RGB images.
Vision-language features extracted from RGB images may have many sources of inconsistency: a painting can be recognized as a wall from a greater distance; a bed object can be misclassified as a sofa at different angles of view; objects on image borders and occluded objects might not be visible enough to provide correct features; inaccurate detection on the object edge can spoil features of the objects behind them; etc.

LiLMaps focuses on incremental implicit language mapping, i.e., when new observations become available incrementally, one-by-one.
This is a typical condition for SLAM, and LiLMaps can be integrated into existing implicit SLAM approaches with minimal changes.
We achieve this with the following key techniques that are presented in this work:

\begin{itemize}
\item \textbf{Adaptive Language Decoder Optimization} dynamically updates the decoder to new discovered language features in the environment providing flexible and sufficient coverage of language representations.

\item \textbf{Measurements Update Strategy} adjusts incoming measurements reusing accumulated and implicitly stored knowledge about the environment to reduce measurements inconsistency.
\end{itemize}

Our experiments show that LiLMaps enables incremental vision-language environment exploration with just a small overhead.

\section{Related Works}
\begin{figure*}[t]
  \centering
  \includegraphics[width=1.0\linewidth]{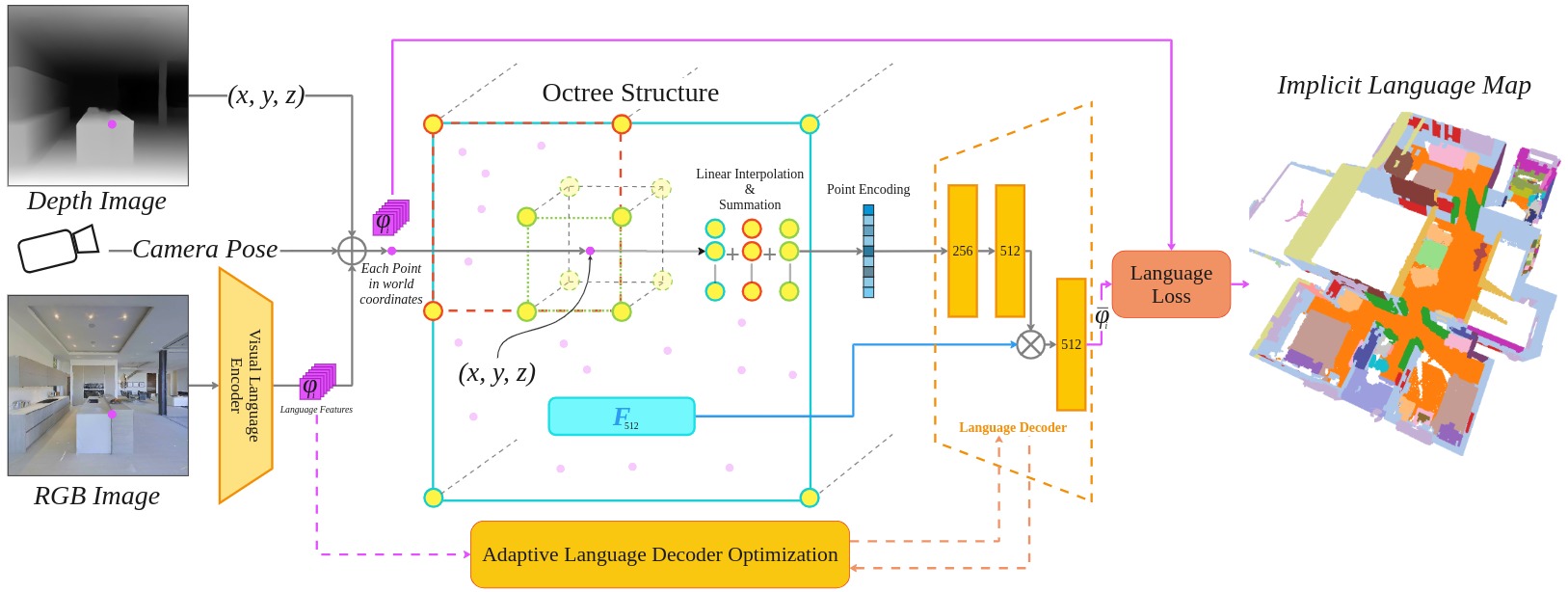}

   \caption{Implicit language mapping. Vision-language features $\varphi$ are extracted from the RGB image. The corresponding points of the depth image are projected to the world coordinate system. Each point can be encoded using its coordinates and octree: the coordinates are used to find the corresponding octree voxels (\textcolor{cyan}{blue}, \textcolor{red}{red}, \textcolor{green}{green}); learnable features stored in the voxels' corners are interpolated and summed, producing the point encoding. $F$ vectors are stored only in the voxels of the coarse octree level (\textcolor{cyan}{blue}). The language decoder reconstructs the language feature $\bar{\varphi}$ in the spatial coordinates of the point based on its encoding and the vector $F$. Language loss optimizes the learnable features and $F$ vectors. After optimization, the language map can be reconstructed in arbitrary spatial coordinates. The language detector is optimized independently of the implicit mapping (\cref{sec:adaptive-optimization}).}
   \label{fig:pipeline}
\end{figure*}
\noindent\textbf{Language and Vision-Language Models.} 
Dynamic execution of natural language commands has been an active research topic for a long time~\cite{anderson2018vision,hong2022bridging}.
Previous works often propose custom environment representations that are difficult to reuse in real applications.
Recently, large language models (LLMs) have received increased attention in this field.
The advantage of LLMs is their ability to be applied to a wide range of tasks.
LLMs can enhance robot abilities to understand and execute natural language commands~\cite{obinata2023foundation,hu2023toward}.
In addition, LLMs have been shown to be successful in guiding object grasping~\cite{dalal2024plan,lynch2023interactive}, navigation~\cite{huang23vlmaps, chen2023open, raman2022planning, ahn2022can, song2023llm} and scene understanding~\cite{ha2022semantic,peng2023openscene,chen2022leveraging}.
Vision-Language Models (VLMs)~\cite{li2022language, ghiasi2022scaling, ranasinghe2022perceptual, mingfengli_seganyclip}, which extract language information about the environment from the provided images, are frequently used in conjunction with LLMs.
For instance, CLIP~\cite{radford2021learning} is one of the most widely used models capable of mapping images into a natural language space.

Often, VLMs transfer only single or batched images into a natural language vector space.
However, some tasks may benefit from mapping the entire environment into the language space.
For example, VLMaps~\cite{huang23vlmaps} suggests enhancing 2D maps with language features and then demonstrates that navigation and detection tasks can be solved directly on these enhanced maps.
Although VLMaps can be used alongside simultaneous localization and mapping (SLAM), its performance depends on localization and mapping quality, and the method is limited to building only 2D language maps.
On the other hand, OpenScene~\cite{peng2023openscene} can generate maps as 3D point clouds with corresponding language features. However, the method performs best in batch-like operations, where all RGB images of the environment, their poses, and 3D point clouds are available in advance, meaning no real-time data is processed.
Another approach, ConceptFusion~\cite{jatavallabhula2023conceptfusion}, demonstrates that language features can be fused into 3D maps using traditional SLAM approaches. SAM3D~\cite{yang2023sam3d} projects SAM segmentation masks~\cite{kirillov2023segment} into 3D and creates 3D scene masks. 

\noindent\textbf{Implicit Representations.}
Implicit representations~\cite{mildenhall2021nerf, reiser2021kilonerf, bloesch2018codeslam, ortiz2022isdf, matsuki2024gaussian} have gained popularity for their compactness and ability to achieve high-resolution reconstructions.
They demonstrate great capabilities in environment mapping.
iMap~\cite{sucar2021imap} uses an RGB-D sensor to perform a real-time SLAM task.
Nice-SLAM~\cite{zhu2022nice} extended the possible sizes of the mapped environment.
SHINE-Mapping~\cite{zhong2023shine} demonstrated implicit mapping of outdoor environments.
Recently, works based on Gaussian splatting~\cite{kerbl20233d} demonstrated exceptional results~\cite{matsuki2024gaussian, zhu2024semgauss, naumann2024nerf, zhu2024loopsplat}.
In addition to geometry, implicit maps demonstrate a successful reconstruction of semantic information~\cite{zhi2021ilabel}, physical properties~\cite{haughton2022real}, and visual features~\cite{mazur2023feature}.

Integration of language features into implicit representation is an actively researched topic.
LERF~\cite{kerr2023lerf} studies the fusion of language features into Radiance Fields, but is limited to small scenes.
LangSplat~\cite{qin2024langsplat} uses Gaussian Splatting~\cite{kerbl20233d} to achieve higher precision and training speed. However, LangSplat demonstrates only the reconstruction of small table-sized scenes and does not consider incremental mapping.

Implicit representations are highly dependent on the type of encoding they use. The encodings employed in the original NeRF work~\cite{mildenhall2021nerf} were able to generalize the predictions~\cite{haughton2022real,mazur2023feature}, but were constrained by the limited size of the environments.
Subsequent works successfully increased training and reconstruction speeds~\cite{reiser2021kilonerf,muller2022instant}.
Gaussian Splatting~\cite{kerbl20233d} is currently one of the most popular encodings due to its speed, simplicity, and high quality reconstruction.
However, some works may benefit from structured encodings such as grid-based~\cite{wang2022go}, octree-based feature volumes~\cite{takikawa2021neural}, or combined ones~\cite{li2024gs}.

Compared to the works discussed above, our approach can build large-scale 3D implicit language maps and can be seamlessly integrated with implicit SLAM methods. 
LiLMaps use a sparse octree-based representation~\cite{takikawa2021neural} to store learnable features, but our method is not tied to any particular representation and can be adapted to others with minimal effort.

\section{Method}\label{sec:method}
We address the task of building an implicit vision-language representation along with environment mapping.
~\cref{sec:architecture} describes our model architecture used in LiLMaps.
During incremental mapping, the future observed objects and their encoded representations are unknown in advance, which makes it challenging to train a language decoder to represent all language features.
To address this issue, in~\cref{sec:adaptive-optimization} we propose the adaptive language decoder optimization strategy that can effectively adjust the decoder to new language features while retaining previously observed ones.
In this work, we employ a visual language encoder but pixel-wise language features are often inconsistent between frames. We address this issue in \cref{sec:measurements-update}.

\subsection{LiLMaps Architecture}\label{sec:architecture}
~\cref{fig:pipeline} shows the architecture of the proposed approach.
Input data for our pipeline are point clouds associated with CLIP language features (language point clouds), as well as camera poses estimated by any external SLAM method. 
We produce language point clouds by extracting language features from an RGB image.
Extracted language features are projected to the point clouds in the world coordinate system using the corresponding depth image and camera pose.
The extraction of language features can be done using per-pixel visual language encoders such as LSeg, OpenSeg, Segment-Anything-CLIP~\cite{mingfengli_seganyclip}, etc. For example, VLMaps~\cite{huang23vlmaps} utilizes LSeg for this purpose.
In this study, the visual language encoder is treated as an external module, and improving its performance is not our focus.

Our goal is to enable implicit vision-language mapping under the conditions of environment exploration when future measurements are not available. 
We use the octree structure as positional encoding to build the implicit representation.
Unless otherwise specified, we consistently use three different levels of the octree to store the features.
Each level of the octree is made up of voxels, and each voxel from a higher level can encompass multiple voxels from lower, more detailed levels.
It should be noted that we use a sparse octree representation, meaning voxels are only present where observations have been made.
When the point clouds are projected to the world coordinates, we find the corresponding voxels in the octree for each point.
Each voxel holds learnable features at its corners. 
These features are shared among voxels that have common corners.

The language features have a high dimensionality and a straightforward solution to encode them in the octree is to increase the size of the learnable features stored in the corners.
However, storing high-dimensional learnable features consumes a significant amount of memory.
Instead, we suggest storing one high-dimensional learnable feature vector $F$ per voxel of the first (coarse) octree level, while keeping the corner features low-dimensional.

To train the implicit representation, we apply the cosine similarity loss between the features decoded from the octree $\bar{\varphi}$ and the vision-language features $\varphi$ of the input point cloud: \vspace*{-2ex}
\begin{equation}
\mathcal{L}_{vl} = -\frac{1}{N}\sum_{i=1}^{N}\textrm{CosineSimilarity}(\varphi_i, \bar{\varphi}_i)\label{eq:vl-loss},
\end{equation}

where $N$ is the number of points with vision-language features in the point cloud.

Note that we encode every new available measurement into the learnable features using \eqref{eq:vl-loss}, but the weights of the language decoder are not updated with this loss.
The optimization of the decoder is described in \cref{sec:adaptive-optimization}.

The decoder reconstructs the language feature $\bar{\varphi}$  in spatial coordinates using feature vector $F$ and corners features of the corresponding voxels.
Before feeding to the decoder, the corner features are linearly interpolated into the reconstruction point and summed across all octree levels.
The decoder consists of three fully connected layers.
The first two layers expand the dimensions of the corner features and produce the element-wise scaling vector for $F$, which is then multiplied by it and passed to the last fully connected layer, which outputs the predicted language feature $\bar{\varphi}$.

\subsection{Adaptive Language Decoder Optimization}\label{sec:adaptive-optimization}
Equation~\eqref{eq:vl-loss} uses our MLP-based language decoder to predict vision-language features $\bar{\varphi}$ based on the encodings stored in the octree. 
However, the new data can contain features that have not been observed before.
In this case, the decoder weights must be updated to be able to reconstruct new features without forgetting the old ones, but re-training of the whole previously mapped environment is computationally expensive.

We propose adaptive language decoder optimization in \cref{alg:decoder-opt}. 
When a new point cloud with vision-language features arrives, we perform decoder optimization before the optimization of octree features described in \cref{sec:architecture}. 

\algdef{SE}[VARIABLES]{Variables}{EndVariables}
   {\algorithmicvariables}
   {\algorithmicend\ \algorithmicvariables}
\algnewcommand{\algorithmicvariables}{\textbf{global names}}

\begin{algorithm}
\caption{Adaptive Language Decoder Optimization}\label{alg:decoder-opt}
\begin{algorithmic}[1]
\Variables
\State $\textrm{LDec}$   \Comment{Language decoder} \label{alg:init-ldecoder}
\State $\textrm{kFeatures}$ $\gets$ $\emptyset$ \Comment{Known features} \label{alg:init-kfeatures}
\State $\textrm{Encodings}$ $\gets$ $\emptyset$  \Comment{Learnable Parameter} \label{alg:init-encodings}
\State $\textrm{FVectors}$ $\gets$ $\emptyset$  \Comment{Learnable Parameter} \label{alg:init-fvectors}
\State $\tau$  \Comment{Cosine similarity threshold}
\State \Call{Loss}{}    \Comment{Cosine similarity loss}
\EndVariables
\Procedure{Optimize}{$\textrm{inFeatures}$}
\State $\textrm{uniqueFeatures} \gets \Call{Unique}{\textrm{inFeatures}, \tau}$ \label{alg:unique}
\State $\textrm{newFeatures} \gets \Call{Unknown}{\textrm{uniqueFeatures},
\newline\hspace*{13.3em}\textrm{kFeatures}, \tau}$ \label{alg:unknown}
\If{$\textrm{len}(\textrm{newFeatures}) = 0$} 
    \State \Return  \Comment{No optimization if no new features}
\EndIf

\State $\textrm{encodings}\gets \Call{Random}{\textrm{len}(\textrm{newFeatures}), m}$ \label{alg:random-enc}

\If{$\textrm{len}(\textrm{FVectors}) = 0$} 
    \State $\textrm{fvectors}\gets \Call{Random}{1, L}$ \label{alg:random-feat}
\Else
    \State $\textrm{fvectors}\gets \Call{Mean}{\textrm{FVectors},\textrm{dim}\mathbin{=}0, 
    \newline\hspace*{11em}\textrm{keepdim}\mathbin{=}\textrm{true}}$ \label{alg:mean-feat}
\EndIf
\State $\textrm{optimizer} \gets \Call{Adam}{[\textrm{encodings}, \textrm{fvectors},\textrm{LDec}]}$\label{alg:init-optimizer}

\State $\textrm{allEncodings} \gets  \textrm{Encodings} \cup \textrm{encodings}$\label{alg:conc1}
\State $\textrm{allFVectors} \gets  \textrm{FVectors} \cup \textrm{fvectors}$\label{alg:conc2}
\State $\textrm{allFeatures} \gets  \textrm{kFeatures} \cup \textrm{newFeatures}$\label{alg:conc3}

\For{$i \gets 0$ to $N_\textrm{opt}$}  
        
        \State {$\bar{\varphi} \gets \textrm{LDec}(\textrm{allEncodings}, \textrm{allFVectors})$}
        \State {$\textrm{fvectors1} \gets \Call{Shuffle}{\textrm{allFVectors}}$}\label{alg:shuffle1}
        \State {$\textrm{loss} \gets \Call{Loss}{\textrm{allFeatures}, \bar{\varphi}}$}\label{alg:loss}
        \State {$\textrm{loss}_{F} \gets \Call{Loss}{\textrm{fvectors1}, \textrm{allFVectors}}$}\label{alg:loss_f}
        \State $\textrm{optimizer.optimize}(\textrm{loss}+\textrm{loss}_{F})$\label{alg:optimize}
\EndFor
\State $\textrm{kFeatures}$ $\gets$ $\textrm{allFeatures} $ \Comment{Update features}\label{alg:update1}
\State $\textrm{Encodings}$ $\gets$ $\textrm{allEncodings}$  \Comment{Update encodings}\label{alg:update2}
\State $\textrm{FVectors}$ $\gets$ $\textrm{allFVectors}$  \Comment{Update FVectors}\label{alg:update3}
\EndProcedure
\end{algorithmic}
\end{algorithm}

The proposed optimization operates with a language decoder (\cref{alg:init-ldecoder}) and learnable parameters.
The learnable parameters are the inputs that are directly forwarded to the decoder.
In our work (\cref{fig:pipeline}), the learnable parameters are $F$ vectors (\cref{alg:init-fvectors}) and the interpolated and summed point encoding (\cref{alg:init-encodings}). 
Note that \cref{alg:decoder-opt} is not limited to our network architecture and can be adapted to other implicit representations by simply replacing the corresponding learnable parameters.

Firstly, we extract only unique language features (\cref{alg:unique}) from all available ones in the input point cloud and then filter out already known features (\cref{alg:unknown}).
In both cases, cosine similarity and a predefined threshold $\tau$ are used to estimate similarity.

For the new features (unobserved features without duplicates), we initialize the learnable parameters: the encodings and $F$ vectors (\cref{alg:random-enc,alg:random-feat,alg:mean-feat}).
Note that the initialized encodings (\cref{alg:random-enc}) correspond to the linearly interpolated and summed features of the corners of the octree (Point Encoding in \cref{fig:pipeline}). 
We initialize only a single $F$ feature vector for all new language features (\cref{alg:random-feat}) because the $F$ feature vector stored in the coarse octree level may be used for points with different language features (\cref{sec:architecture}).
We also regularize new $F$ vectors by enforcing them to be similar to existing ones (\cref{alg:shuffle1,alg:loss_f}). 
However, this regularization is optional and can be omitted or changed to any other regularization required by the corresponding implicit representation. 

The proposed adaptive optimization approach optimizes the decoder for unobserved language features and finds the learnable parameters for them.
Only the decoder and new learnable parameters are optimized (\cref{alg:init-optimizer}).
The already known features and their encodings are not optimized, but used to prevent forgetting (\cref{alg:conc1,alg:conc2,alg:conc3}). 
After optimization, we update the list of known vision-language features and their learnable parameters (\cref{alg:update1,alg:update2,alg:update3}).

The proposed optimization efficiently stores only a small number of known features for replay (\cref{alg:init-kfeatures}), as demonstrated in experiments (\cref{fig:barplot}).
It enables fast decoder optimization using vectorization.
Moreover, the language decoder and the learnable parameters are optimized only if new language features are observed.

\begin{table}
  \centering
  {\small{
  \begin{tabular}{@{}lll@{}}
    \toprule
    Parameter & Symbol & Value \\
    \midrule
    Similarity threshold & $\tau$ & 0.02\\
    Used octree levels &   & 8,9,10\\
    Fine level resolution &   & 0.05 [m]\\
    Learnable features size & $m$ & 16 \\
    F vectors size & $L$ & 512\\
    Iterations per decoder optimization & $N_{\textrm{opt}}$ & 100 \\
    Iterations per mapping loss~\cref{eq:vl-loss} &   & 100 \\
    \bottomrule
  \end{tabular}
  }}
  \caption{LiLMaps parameters and their values in the experiments.}
  \label{tab:parameters}
\end{table}

\subsection{Measurement Update Strategy}\label{sec:measurements-update}
During incremental mapping, new observations added to the map should not corrupt previous measurements. 
However, vision-language features predicted by the visual encoder may not be consistent between frames.
VLMaps~\cite{huang23vlmaps} averaged the language features of the objects received from different views.
We note that the averaging can be done in a recursive form:
\begin{equation}
\mathcal{A}_{n} = \frac{1}{N}\sum_{i=1}^{N}\varphi_i \;\;\;= \frac{n-1}{n}\mathcal{A}_{n-1} + \frac{\varphi_n}{n}, \textrm{with } \mathcal{A}_{0}\!=\!0.
\end{equation}
In this work, we propose to use as target for training in \cref{eq:vl-loss} a weighted average $\varphi_n^*$ between observations $\varphi_n$ and the features $\bar{\varphi}_{n-1}$ already stored in the map:
\begin{equation}\label{mean-update}
{\varphi}_n^* =  \frac{n-1}{n}\bar{\varphi}_{n-1} + \frac{\varphi_n}{n}, \;\;\; \textrm{with } \bar{\varphi}_0\!=\!0.
\end{equation}
This averaging is especially useful for noisy measurements such as vision-language features because they may significantly vary with distance to the objects or point of view (\cref{exp:mapping-quality}).
Mapping with \cref{mean-update} forces the map to store all observations similarly to~\cite{huang23vlmaps}.
However, we observed better results when not all previous data are stored and decided to use exponential smoothing instead of averaging:
\begin{equation}
{\varphi}_{n}^* =  \alpha \bar{\varphi}_{n-1} + (1\!-\!\alpha)\varphi_n\;\;\; \textrm{with } \bar{\varphi}_0\!=\!0,
\end{equation}
where $\alpha$ is set dynamically to higher values if new measurements $\varphi_n$ are more different from the previously optimized map features $\bar{\varphi}_{n-1}$ and lower otherwise:
\begin{equation}
\alpha = \frac{\textrm{CosineSimilarity}(\varphi_i, \bar{\varphi}_i)}{0.5 + \textrm{CosineSimilarity}(\varphi_i, \bar{\varphi}_i)}.
\end{equation}



\section{Experiments}

\renewcommand{\arraystretch}{1.2}
\setlength{\tabcolsep}{2pt}
\begin{table*}[htbp]
\begin{center}
\begin{tabular}{l| cccc|cccc|cccc|cccc|cccc}
\hline
\multirow{2}{*}{\backslashbox{\textbf{Approach}}{\textbf{Sequence}}} &
\multicolumn{4}{c|}{5LpN3gDmAk7\_1}  & 
\multicolumn{4}{c|}{YmJkqBEsHnH\_1} &
\multicolumn{4}{c|}{gTV8FGcVJC9\_1}  &
\multicolumn{4}{c|}{jh4fc5c5qoQ\_1} &
\multicolumn{4}{c}{JmbYfDe2QKZ\_2}\\
& 
\normalsize{\textbf{A}} & \normalsize{\textbf{mR}} & \normalsize{\textbf{mP}} & \normalsize{\textbf{mIoU}} &
\normalsize{\textbf{A}} & \normalsize{\textbf{mR}} & \normalsize{\textbf{mP}} & \normalsize{\textbf{mIoU}} &
\normalsize{\textbf{A}} & \normalsize{\textbf{mR}} & \normalsize{\textbf{mP}} & \normalsize{\textbf{mIoU}}&
\normalsize{\textbf{A}} & \normalsize{\textbf{mR}} & \normalsize{\textbf{mP}} & \normalsize{\textbf{mIoU}}&
\normalsize{\textbf{A}} & \normalsize{\textbf{mR}} & \normalsize{\textbf{mP}} & \normalsize{\textbf{mIoU}}\\

\hline

LiLMaps*$_\textrm{GT}$ & 
97 & 97 & 96 & 93 &
98 & 93 & 92 & 89 & 
98 & 98 & 95 & 94 &
98 & 96 & 88 & 85 &
95 & 96 & 92 & 89 \\
LiLMaps$_\textrm{GT}$ & 
97 & 97 & 93 & 91 & 
96 & 94 & 96 & 90 & 
97 & 98 & 93 & 92 &
98 & 98 & 92 & 89 &
 95 & 96 & 90 & 87 \\
 \hline
LiLMaps*$_\textrm{SEM}$ & 
84 & 77 & 70 & 57 & 
84 & 77 & 82 & 65 &
85 & 85 & 84 & 73 &
83 & 78 & 73 & 61 &
78 & 77 & 77 & 63 \\
LiLMaps$_\textrm{SEM}$ & 
\textbf{88} & \textbf{86} & \textbf{75} & \textbf{66} &
\textbf{90} & \textbf{82} & \textbf{87} & \textbf{72} &
\textbf{90} & \textbf{90} & \textbf{86} & \textbf{79} &
\textbf{90} & \textbf{84} & \textbf{82} & \textbf{71} &
\textbf{85} & \textbf{88} & \textbf{82} & \textbf{74} \\
OpenScene~\cite{peng2023openscene} & 
68 & 45 & 67 & 36 & 
63 & 50 & 77 & 41 & 
61 & 49 & 60 & 36 &
77 & 52 & 59 & 39 &
56 & 51 & 67 & 41 \\
\hline
LiLMaps*$_\textrm{LSeg}$ & 
64 & 29 & 46 & 21 & 
52 & 44 & 58 & 31 & 
65 & 48 & 59 & 32 &
59 & 32 & 44 & 21 &
53 & 41 & 50 & 33 \\
LiLMaps$_\textrm{LSeg}$ & 
68 & 37 & 57 & 26 & 
57 & 56 & 60 & 34 & 
70 & 50 & 63 & 33 &
73 & 39 & 58 & 27 &
56 & 42 & 56 & 34 \\

VLMaps$^\textrm{mean}_\textrm{2D}$ & 
28 & - & - & 19 & 
28 & - & - & 19 & 
28 & - & - & 19 &
28 & - & - & 19 &
28 & - & - & 19 \\
\hline
\end{tabular}
\end{center}
\caption{Language mapping quality evaluation: accuracy (A),  recall (mR), precision (mP) and mean IoU (mIoU) in [\%].}\label{tab:map_quality}
\end{table*}
\renewcommand{\arraystretch}{1}
In the experiments, we validate that our method can be used for incremental implicit mapping of language features.
We use depth, semantic, and RGB images provided by~\cite{huang23vlmaps} through the Habitat simulator using Matterport3D~\cite{Matterport3D}.
Matterport3D provides ground truth meshes with each face assigned to a class label.
To get ground truth point clouds with the corresponding language features, we sample the meshes and encode their labels using CLIP\cite{radford2021learning}. 
During all experiments, we project depth images into 3D using the current camera pose to obtain input point clouds.
The parameters we use during the experiments are summarized in \cref{tab:parameters}.
\begin{figure}[t]
  \centering
  \includegraphics[width=1.0\linewidth]{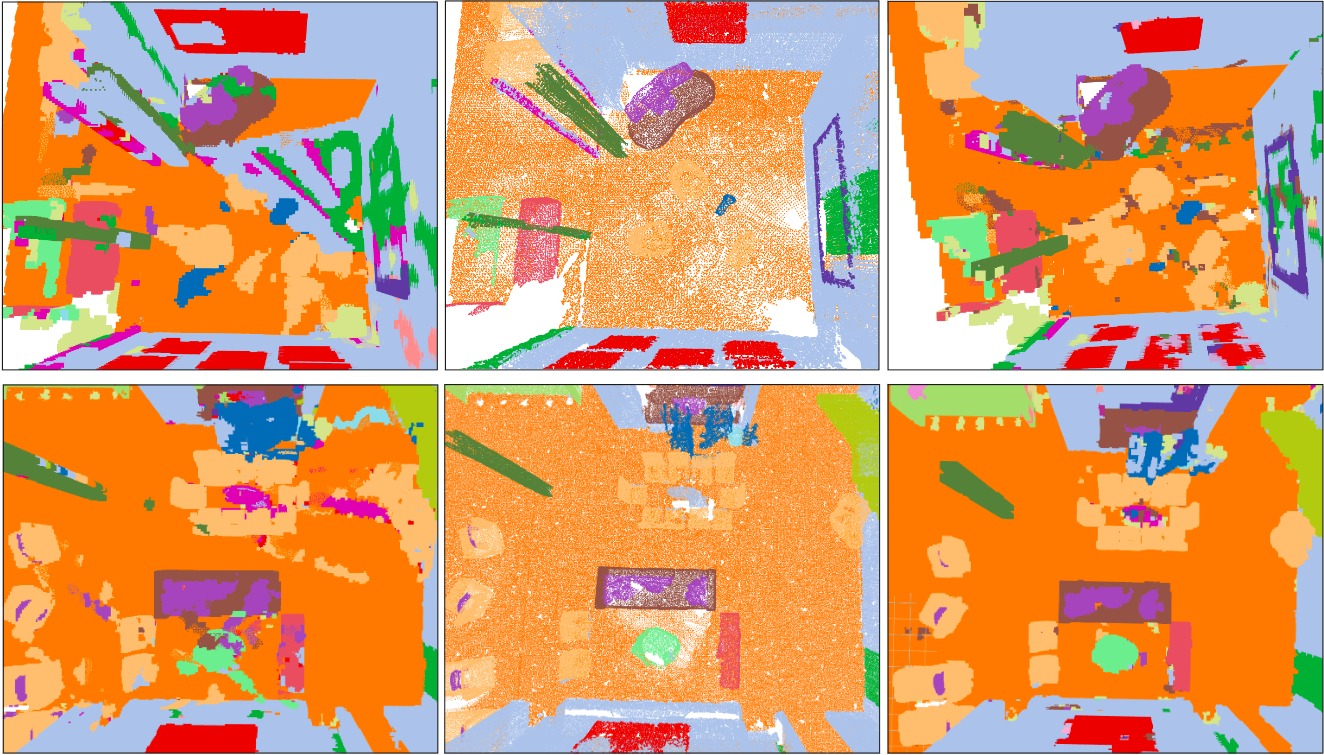}

   \caption{\textit{Left:} Environments reconstructed without measurement update; \textit{Middle:} Ground Truth; \textit{Right:} Environments reconstructed with measurement update.}
   \label{fig:measurements-update}
\end{figure}

\subsection{Mapping Quality}\label{exp:mapping-quality}
We evaluate accuracy, recall, precision, and intersection over union for our implicit language map. Accuracy is defined as the number of points with correctly reconstructed language features divided by the total number of points.
The values are compared with the OpenScene 3D model~\cite{peng2023openscene} trained on Matterport3D~\cite{Matterport3D}.
The results of random sequences are presented in \cref{tab:map_quality}.
To validate the measurement update strategy, we also present results with deactivated measurement update, marked by an asterisk (LiLMaps*).

\begin{figure}[t]
  \centering
  \includegraphics[width=1.0\linewidth]{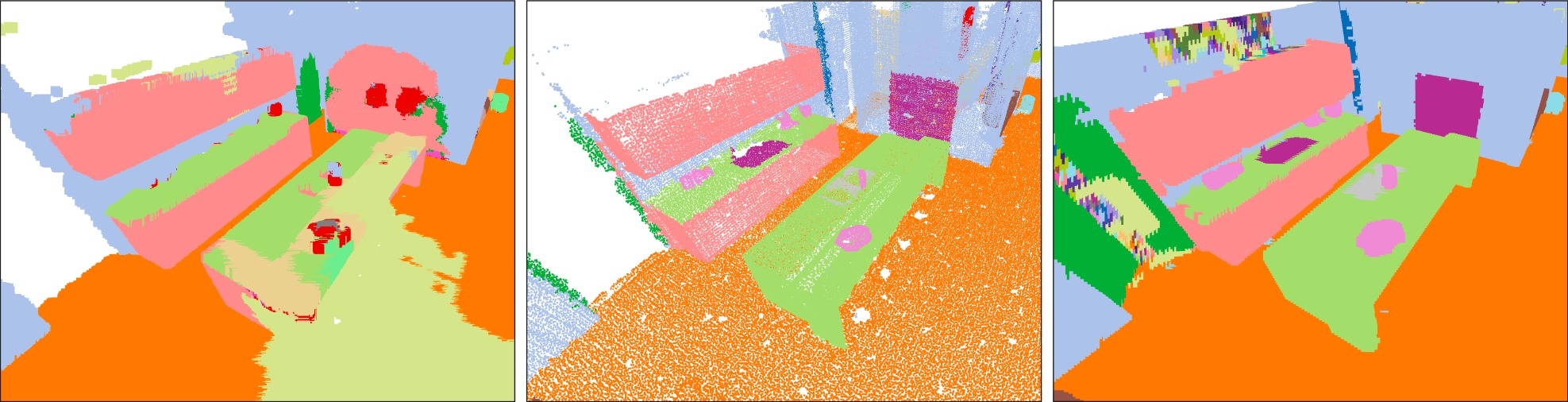}

   \caption{\textit{Left:} Language map produced by OpenScene 3D~\cite{peng2023openscene};
   \textit{Middle:} Ground Truth; \textit{Right:} Language map created by LiLMaps.}
   \label{fig:openscene}
\end{figure}

For LiLMaps$_\textrm{GT}$ and LiLMaps*$_\textrm{GT}$, the language features of a measurement are obtained directly from the closest points of the ground truth point cloud. 
This allows us to estimate upper-bound performance with close-to-ideal input data. 
However, simulation measurements and ground truth points sampled from uneven meshes do not always match perfectly.
As a result, some input points may have wrong language features or do not have language features at all.

LiLMaps$_\textrm{SEM}$ and LiLMaps*$_\textrm{SEM}$ denote experiments in which language features are extracted from semantic images.
Simulated semantic images have incorrect labels due to mesh discontinuities and on object edges.
This allows us to test LiLMaps when the input data are closer to the real ones, e.g. when the data are imprecise and inconsistent between frames.

Our approach demonstrates the best performance when used with the $\textrm{GT}$ data. 
The introduction of the Measurements Update does not change the performance significantly, as the input features are precise and consistent between frames in this case.
\renewcommand{\arraystretch}{1.5}
\setlength{\tabcolsep}{4pt}
\begin{table*}[htbp]

\begin{center}
\begin{tabular}{l| ccccc}
\hline
\multirow{2}{*}{\textbf{Decoders}} & \multicolumn{5}{c}{\textbf{F1-Score}}\\
&

\textbf{$\uparrow$100\% -- 90\%} & \textbf{$\downarrow$90\% -- 80\%} &  \textbf{$\downarrow\downarrow$80\% -- 70\%}  & \textbf{$\downarrow\downarrow\downarrow$70\% -- 50\%} &\textbf{$\downarrow\downarrow\downarrow\downarrow$50\% -- 0\%}\\
\hline

LiLMaps$^\textrm{simple}_\textrm{Adaptive}$& 
20 & 2 & 1 (picture) & 0 & 0 \\
LiLMaps$^\textrm{simple}_\textrm{Pretrained}$& 
20 & 2 & 0 & 1 (towel) & 0\\
OpenScene$^\textrm{MT}_\textrm{HEAD}$ & 
16 & 4 & 2 & 0 & 1 (objects) \\

OpenScene$^\textrm{SC}_\textrm{HEAD}$ & 
16 & 5 & 1 & 0 & 1 (objects) \\

OpenScene$^\textrm{NS}_\textrm{HEAD}$ & 
9 & 8 & 1 & 3 & 2 (picture, shelving) \\

\hline
\end{tabular}
\end{center} \vspace*{-2ex}
\caption{Number of classes falling into different F1-score ranges for compared decoders.
A method is considered to be better if a larger number of classes are listed in the first column with $\textrm{F1}\!>\!90\%$.}\label{tab:decoder_optimization_eval}
\end{table*}
\renewcommand{\arraystretch}{1}

\begin{figure}[t]
  \centering
  \includegraphics[width=1.0\linewidth]{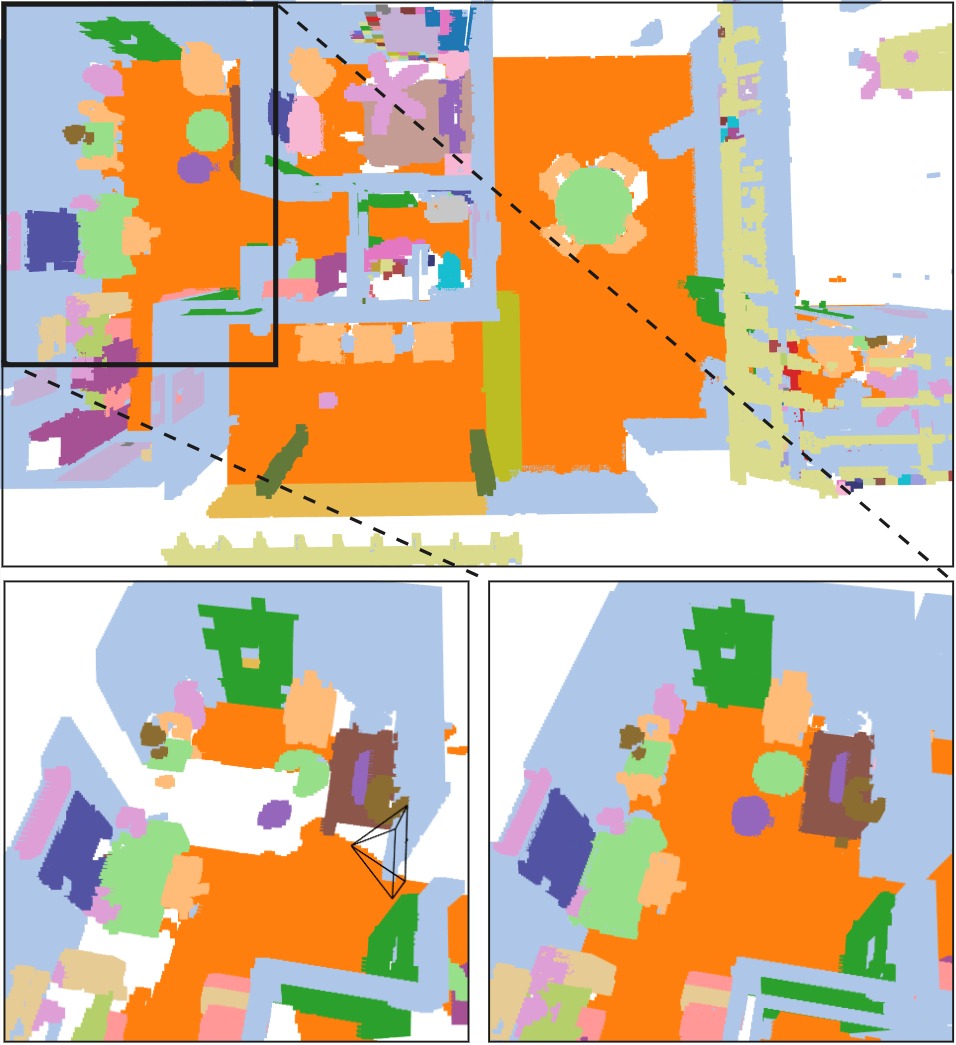}

   \caption{Language map incrementally created with our adaptive optimization. \textit{Bottom Left:} A region mapped in the beginning. \textit{Bottom Right:} The same region after the mapping is completed. All initially mapped objects remain unchanged.}
   \label{fig:incremental-learning}
\end{figure}

As expected, LiLMaps$_\textrm{SEM}$ and LiLMaps*$_\textrm{SEM}$ yield worse results due to the inconsistency of the input data, but both still outperform OpenScene~\cite{peng2023openscene}.
LiLMaps$_\textrm{SEM}$ outperforms LiLMaps*$_\textrm{SEM}$ due to the proposed measurement update technique which addresses potential data inconsistencies. \cref{fig:measurements-update} shows maps learned with activated and deactivated measurement update procedure.
Enabling measurements update results in a cleaner final map, which is crucial for object detection and navigation.

\cref{fig:openscene} compares a LiLMaps reconstruction with the prediction of the OpenScene 3D model. Despite OpenScene being trained on the same dataset, it struggles with certain labels and completely misses labels such as "TV monitor", "appliances", "stool", whereas our approach achieves high accuracy for these labels and does not completely miss objects.

\cref{tab:map_quality} compares our approach combined with the LSeg model (LiLMaps$_\textrm{LSeg}$ and LiLMaps*$_\textrm{LSeg}$) and VLMaps (VLMaps$^\textrm{mean}_\textrm{2D}$) with metrics reported in~\cite{huang23vlmaps}.
LSeg frequently misses objects (e.g., segments a painting as a wall) or provides wrong language features (e.g., detects a bed as a sofa), which significantly influence the final results.
Improving the quality of per-pixel language segmentation is beyond the scope of this research, however.
In all cases, our 3D reconstructed language maps yield better results than the mean results reported in VLMaps~\cite{huang23vlmaps} for their 2D maps.

\subsection{Adaptive Language Decoder Optimization}

We demonstrate the impact of our Adaptive Optimization Strategy on sequence 5LpN3gDmAk7\_1 with $\textrm{GT}$ labels.
We compare the performance of the decoder trained with our Adaptive Optimization with other decoders built from pre-trained models.
We chose OpenScene~\cite{peng2023openscene} 3D model's head as the pre-trained decoder because it can predict language features for arbitrary input point clouds.
For fair comparison, we changed our language decoder architecture (LiLMaps$^\textrm{simple}$) to match the architecture of OpenScene's head.
This head does not allow us to use the learnable vectors $F$, however, and therefore the results in \cref{tab:decoder_optimization_eval} are presented when only the learnable corner features with $m\!=\!96$ are optimized.

\cref{tab:decoder_optimization_eval} summarizes number of classes with distinct F1-score qualities using different optimization types: decoder trained with the proposed adaptive optimization (LiLMaps$^\textrm{simple}_\textrm{adaptive}$); decoder pre-trained with the proposed optimization (LiLMaps$^\textrm{simple}_\textrm{pretraiend}$); and pre-trained and fixed headings of OpenScene~\cite{peng2023openscene}, trained on different datasets (Matterport3D~\cite{Matterport3D}, nuScenes~\cite{caesar2020nuscenes}, ScanNet~\cite{dai2017scannet}).

\begin{figure}[t]
  \centering
  \includegraphics[width=1.0\linewidth]{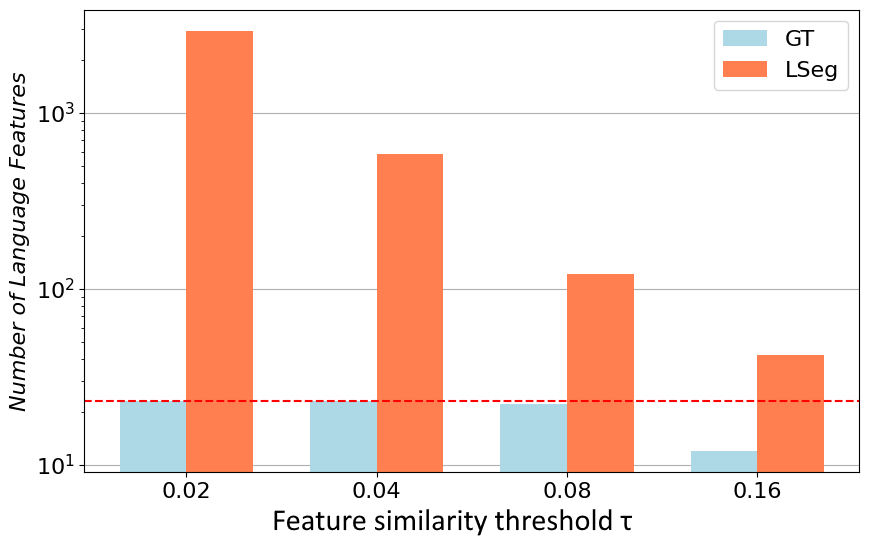}

   \caption{Number of language features stored for the adaptive optimization with varying  feature similarity thresholds $\tau$. \textit{Blue:} Language features are extracted from ground truth (GT) data; \textit{Orange:} Language features are extracted by LSeg; \textit{Red line:} Total number of different GT classes presented in the scene.}
   \label{fig:barplot}
\end{figure}

To obtain LiLMaps$^\textrm{simple}_\textrm{pretraiend}$ we extract all available labels from the scene, convert them to language features using CLIP and use our adaptive language decoder optimization with all features at once.
The additional possibility of pre-training the decoder in advance without any real measurements may be useful for some applications.
Our adaptive language decoder optimization allows to adjust pre-trained decoders online if necessary, but for this experiment, we do not update pre-trained models (LiLMaps$^\textrm{simple}_\textrm{pretraiend}$, OpenScene$^\textrm{MT}_\textrm{HEAD}$, OpenScene$^\textrm{SC}_\textrm{HEAD}$, OpenScene$^\textrm{NS}_\textrm{HEAD}$) during the mapping.

The final results of LiLMaps$^\textrm{simple}_\textrm{adaptive}$ are similar to those of LiLMaps$^\textrm{simple}_\textrm{pretraiend}$ because every time a new object is observed, the corresponding language features are included in the Adaptive Optimization and the decoder of LiLMaps$^\textrm{simple}_\textrm{adaptive}$ is updated to represent new features without forgetting the old ones.
Adaptive and pre-trained LiLMaps models may have minor differences in the results due to their initial states and optimization processes being different.
In \cref{fig:incremental-learning} we demonstrate that the proposed adaptive optimization can incrementally extend the decoder to represent new features without catastrophic forgetting of language features observed in the beginning.

The results of \cref{tab:decoder_optimization_eval} show that the proposed adaptive optimization LiLMaps$^{simple}_{Adaptive}$ performs better than pre-trained and fixed models.
Our adaptive optimization fits the model to a specific scene while pre-trained decoders (in this case OpenScene's heads) are trained for general language prediction.
If a model trained for general prediction is used, then some language features of the environment may be poorly represented in it (e.g. picture and shelving in \cref{tab:decoder_optimization_eval}), while well-represented features may be irrelevant for the specific scene.
This can be seen in the results of OpenScene$^\textrm{NS}_\textrm{HEAD}$.
OpenScene$^\textrm{MT}_\textrm{HEAD}$ is trained on Matterport3D~\cite{Matterport3D} and has better results.
OpenScene$^\textrm{SC}_\textrm{HEAD}$ is trained on ScanNet~\cite{dai2017scannet} which is similar to Matterport3D that explains the similar results.
However, OpenScene$^\textrm{NS}_\textrm{HEAD}$ pre-trained on NuScenes~\cite{caesar2020nuscenes} has significant degradation in the results because the NuScenes environment is more different from Matterport3D.
Moreover, our adaptive language decoder optimization allows one to build a custom decoder architecture while employing pre-trained models could restrict available architecture options.

We analyze different values of the threshold $\tau$ used to extract unique and unknown features from all input features.
\cref{fig:barplot} shows the final number of features that were considered distinguished and were involved in the optimization at the end of mapping. 
Lower $\tau$ values lead to a larger number of features, but they are still memory efficient.
For comparison, the stored features are collected from hundreds of high-resolution images, but their final number is less than $0.5\%$ of the number of pixels in a single image with resolution 640$\times$480. 
During all tests, adaptive language decoder optimization operated at a rate of 4 frames per second (fps).
To achieve real-time performance, adaptive optimization can be executed in parallel to mapping. 


\section{Conclusion}
\begin{figure}[t]
  \centering
  \includegraphics[width=1.0\linewidth]{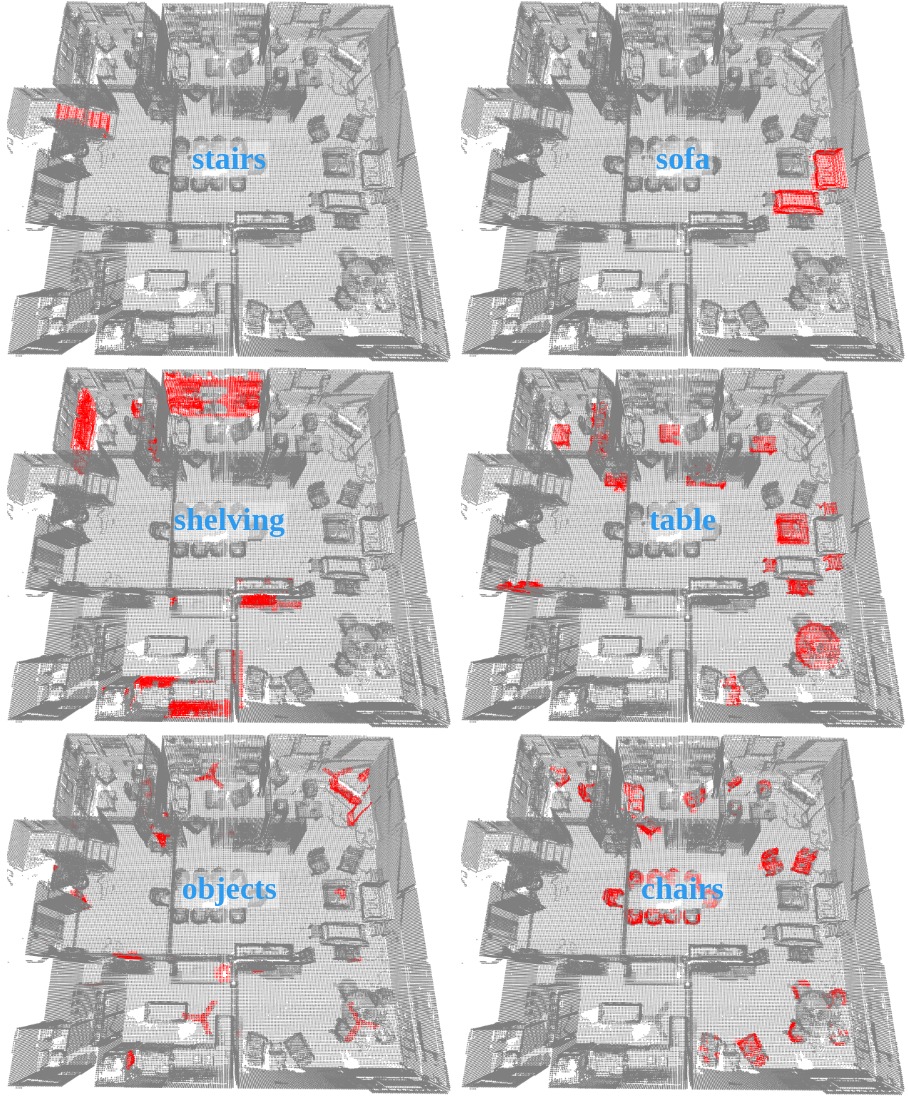}

   \caption{3D language-based object detection performed on our language map. LiLMaps creates an implicit language map which is reconstructed and queried for different objects. The \textcolor{red}{highest correspondences} between the reconstructed language map and corresponding request (\textcolor{cyan}{blue}) are highlighted in red.}
   \label{fig:detections}
\end{figure}

In this work, we presented an implicit language mapping approach called LiLMaps.
We address the problem of unseen language features that appear during the mapping process and the problem of inconsistencies between frames.
Currently, LiLMaps is the only approach capable of large-scale incremental implicit language mapping.
It can be used alone and enables a variety of interactions with the environment, for instance, 3D language-based object detection (\cref{fig:detections}). Additionally, it can be integrated into existing implicit mapping approaches, introducing only slight overhead.

We evaluated LiLMaps on the public dataset commonly used in related works.
Based on the results, we outperform similar works in terms of language mapping quality.
However, LiLMaps significantly depends on the quality of visual language features produced by the encoder, which is considered to be an external module in our study.
The importance of this dependency is reduced by the proposed Measurement Update strategy that handles inconsistency between frames.
We demonstrated that LiLMaps can adapt to the environment outperforming pre-trained decoders.
The proposed Adaptive Optimization demonstrates the ability to prepare decoders given arbitrary language features without the need for actual observations.

\section*{Acknowledgment}
This research was funded by the German Federal Ministry of Education and Research (BMBF) in the project WestAI -- AI Service Center West, grant no. 01IS22094A.

{\small
\bibliographystyle{ieee_fullname}
\bibliography{main.bib}

\begin{thebibliography}{10}\itemsep=-1pt

\bibitem{ahn2022can}
Michael Ahn, Anthony Brohan, Noah Brown, Yevgen Chebotar, Omar Cortes, Byron David, Chelsea Finn, Chuyuan Fu, Keerthana Gopalakrishnan, Karol Hausman, et~al.
\newblock Do as {I} can, not as {I} say: Grounding language in robotic affordances.
\newblock {\em arXiv preprint arXiv:2204.01691}, 2022.

\bibitem{anderson2018vision}
Peter Anderson, Qi Wu, Damien Teney, Jake Bruce, Mark Johnson, Niko S{\"u}nderhauf, Ian Reid, Stephen Gould, and Anton Van Den~Hengel.
\newblock Vision-and-language navigation: Interpreting visually-grounded navigation instructions in real environments.
\newblock In {\em IEEE Conference on Computer Vision and Pattern Recognition (CVPR)}, pages 3674--3683, 2018.

\bibitem{bloesch2018codeslam}
Michael Bloesch, Jan Czarnowski, Ronald Clark, Stefan Leutenegger, and Andrew~J Davison.
\newblock {CodeSLAM} -- learning a compact, optimisable representation for dense visual {SLAM}.
\newblock In {\em IEEE Conference on Computer Vision and Pattern Recognition (CVPR)}, pages 2560--2568, 2018.

\bibitem{caesar2020nuscenes}
Holger Caesar, Varun Bankiti, Alex~H Lang, Sourabh Vora, Venice~Erin Liong, Qiang Xu, Anush Krishnan, Yu Pan, Giancarlo Baldan, and Oscar Beijbom.
\newblock {nuScenes}: A multimodal dataset for autonomous driving.
\newblock In {\em IEEE/CVF Conference on Computer Vision and Pattern Recognition (CVPR)}, pages 11621--11631, 2020.

\bibitem{Matterport3D}
Angel Chang, Angela Dai, Thomas Funkhouser, Maciej Halber, Matthias Niessner, Manolis Savva, Shuran Song, Andy Zeng, and Yinda Zhang.
\newblock {Matterport3D}: Learning from rgb-d data in indoor environments.
\newblock {\em International Conference on 3D Vision (3DV)}, pages 667--676, 2017.

\bibitem{chen2023open}
Boyuan Chen, Fei Xia, Brian Ichter, Kanishka Rao, Keerthana Gopalakrishnan, Michael~S Ryoo, Austin Stone, and Daniel Kappler.
\newblock Open-vocabulary queryable scene representations for real world planning.
\newblock In {\em IEEE International Conference on Robotics and Automation (ICRA)}, pages 11509--11522, 2023.

\bibitem{chen2022leveraging}
William Chen, Siyi Hu, Rajat Talak, and Luca Carlone.
\newblock Leveraging large language models for robot {3D} scene understanding.
\newblock {\em arXiv preprint arXiv:2209.05629}, 2022.

\bibitem{dai2017scannet}
Angela Dai, Angel~X. Chang, Manolis Savva, Maciej Halber, Thomas~A. Funkhouser, and Matthias Nie{\ss}ner.
\newblock {ScanNet}: Richly-annotated 3d reconstructions of indoor scenes.
\newblock In {\em IEEE Conference on Computer Vision and Pattern Recognition (CVPR)}, pages 2432--2443, 2017.

\bibitem{dalal2024plan}
Murtaza Dalal, Tarun Chiruvolu, Devendra Chaplot, and Ruslan Salakhutdinov.
\newblock {Plan-Seq-Learn}: Language model guided {RL} for solving long horizon robotics tasks.
\newblock In {\em 12th International Conference on Learning Representations (ICLR)}, 2024.

\bibitem{ghiasi2022scaling}
Golnaz Ghiasi, Xiuye Gu, Yin Cui, and Tsung-Yi Lin.
\newblock Scaling open-vocabulary image segmentation with image-level labels.
\newblock In {\em European Conference on Computer Vision (ECCV)}, pages 540--557. Springer, 2022.

\bibitem{ha2022semantic}
Huy Ha and Shuran Song.
\newblock {Semantic Abstraction}: Open-world {3D} scene understanding from {2D} vision-language models.
\newblock In {\em Conference on Robot Learning (CoRL)}, volume 205 of {\em Proceedings of Machine Learning Research}, pages 643--653. {PMLR}, 2022.

\bibitem{haughton2022real}
Iain Haughton, Edgar Sucar, Andr{\'{e}} Mouton, Edward Johns, and Andrew~J. Davison.
\newblock Real-time mapping of physical scene properties with an autonomous robot experimenter.
\newblock In {\em Conference on Robot Learning (CoRL)}, volume 205 of {\em Proceedings of Machine Learning Research}, pages 118--127. {PMLR}, 2022.

\bibitem{hong2022bridging}
Yicong Hong, Zun Wang, Qi Wu, and Stephen Gould.
\newblock Bridging the gap between learning in discrete and continuous environments for vision-and-language navigation.
\newblock In {\em IEEE/CVF Conference on Computer Vision and Pattern Recognition (CVPR)}, pages 15439--15449, 2022.

\bibitem{hu2023toward}
Yafei Hu, Quanting Xie, Vidhi Jain, Jonathan Francis, Jay Patrikar, Nikhil Keetha, Seungchan Kim, Yaqi Xie, Tianyi Zhang, Zhibo Zhao, et~al.
\newblock Toward general-purpose robots via foundation models: A survey and meta-analysis.
\newblock {\em arXiv preprint arXiv:2312.08782}, 2023.

\bibitem{huang23vlmaps}
Chenguang Huang, Oier Mees, Andy Zeng, and Wolfram Burgard.
\newblock Visual language maps for robot navigation.
\newblock In {\em IEEE International Conference on Robotics and Automation (ICRA)}, 2023.

\bibitem{jatavallabhula2023conceptfusion}
Krishna~Murthy Jatavallabhula, Alihusein Kuwajerwala, Qiao Gu, Mohd. Omama, Ganesh Iyer, Soroush Saryazdi, Tao Chen, Alaa Maalouf, Shuang Li, Nikhil~Varma Keetha, Ayush Tewari, Joshua~B. Tenenbaum, Celso~Miguel de Melo, K.~Madhava Krishna, Liam Paull, Florian Shkurti, and Antonio Torralba.
\newblock {ConceptFusion}: Open-set multimodal {3D} mapping.
\newblock In {\em Robotics: Science and Systems XIX (RSS)}, 2023.

\bibitem{kerbl20233d}
Bernhard Kerbl, Georgios Kopanas, Thomas Leimk{\"u}hler, and George Drettakis.
\newblock {3D} {Gaussian} splatting for real-time radiance field rendering.
\newblock {\em ACM Transactions on Graphics (TOG)}, 42(4):139--1, 2023.

\bibitem{kerr2023lerf}
Justin Kerr, Chung~Min Kim, Ken Goldberg, Angjoo Kanazawa, and Matthew Tancik.
\newblock {LERF}: Language embedded radiance fields.
\newblock In {\em IEEE/CVF International Conference on Computer Vision (ICCV)}, pages 19729--19739, 2023.

\bibitem{kirillov2023segment}
Alexander Kirillov, Eric Mintun, Nikhila Ravi, Hanzi Mao, Chloe Rolland, Laura Gustafson, Tete Xiao, Spencer Whitehead, Alexander~C Berg, Wan-Yen Lo, et~al.
\newblock Segment anything.
\newblock In {\em IEEE/CVF International Conference on Computer Vision (ICCV)}, pages 4015--4026, 2023.

\bibitem{li2022language}
Boyi Li, Kilian~Q. Weinberger, Serge~J. Belongie, Vladlen Koltun, and Ren{\'{e}} Ranftl.
\newblock Language-driven semantic segmentation.
\newblock In {\em 10th International Conference on Learning Representations (ICLR)}, 2022.

\bibitem{li2024gs}
Jiaze Li, Zhengyu Wen, Luo Zhang, Jiangbei Hu, Fei Hou, Zhebin Zhang, and Ying He.
\newblock {GS-Octree}: Octree-based {3D} {Gaussian} splatting for robust object-level {3D} reconstruction under strong lighting.
\newblock {\em Computer Graphics Forum (CGF)}, 43(7):i--xxii, 2024.

\bibitem{mingfengli_seganyclip}
Ming-Feng Li.
\newblock Per-pixel features: Mating segment-anything with {CLIP}, 2023.

\bibitem{lynch2023interactive}
Corey Lynch, Ayzaan Wahid, Jonathan Tompson, Tianli Ding, James Betker, Robert Baruch, Travis Armstrong, and Pete Florence.
\newblock Interactive language: Talking to robots in real time.
\newblock {\em IEEE Robotics and Automation Letters (RA-L)}, 2023.

\bibitem{matsuki2024gaussian}
Hidenobu Matsuki, Riku Murai, Paul~HJ Kelly, and Andrew~J Davison.
\newblock Gaussian splatting {SLAM}.
\newblock In {\em IEEE/CVF Conference on Computer Vision and Pattern Recognition (CVPR)}, pages 18039--18048, 2024.

\bibitem{mazur2023feature}
Kirill Mazur, Edgar Sucar, and Andrew~J Davison.
\newblock Feature-realistic neural fusion for real-time, open set scene understanding.
\newblock In {\em IEEE International Conference on Robotics and Automation (ICRA)}, pages 8201--8207, 2023.

\bibitem{mildenhall2021nerf}
Ben Mildenhall, Pratul~P Srinivasan, Matthew Tancik, Jonathan~T Barron, Ravi Ramamoorthi, and Ren Ng.
\newblock {NeRF}: Representing scenes as neural radiance fields for view synthesis.
\newblock {\em Communications of the ACM}, 65(1):99--106, 2021.

\bibitem{muller2022instant}
Thomas M{\"u}ller, Alex Evans, Christoph Schied, and Alexander Keller.
\newblock Instant neural graphics primitives with a multiresolution hash encoding.
\newblock {\em ACM Transactions on Graphics (TOG)}, 41(4):1--15, 2022.

\bibitem{naumann2024nerf}
Jens Naumann, Binbin Xu, Stefan Leutenegger, and Xingxing Zuo.
\newblock {NeRF-VO}: Real-time sparse visual odometry with neural radiance fields.
\newblock {\em IEEE Robotics and Automation Letters (RA-L)}, 9(8):7278--7285, 2024.

\bibitem{obinata2023foundation}
Yoshiki Obinata, Naoaki Kanazawa, Kento Kawaharazuka, Iori Yanokura, Soonhyo Kim, Kei Okada, and Masayuki Inaba.
\newblock Foundation model based open vocabulary task planning and executive system for general purpose service robots.
\newblock {\em arXiv preprint arXiv:2308.03357}, 2023.

\bibitem{ortiz2022isdf}
Joseph Ortiz, Alexander Clegg, Jing Dong, Edgar Sucar, David Novotn{\'{y}}, Michael Zollh{\"{o}}fer, and Mustafa Mukadam.
\newblock {iSDF}: Real-time neural signed distance fields for robot perception.
\newblock In {\em Robotics: Science and Systems XVIII (RSS)}, 2022.

\bibitem{peng2023openscene}
Songyou Peng, Kyle Genova, Chiyu Jiang, Andrea Tagliasacchi, Marc Pollefeys, Thomas Funkhouser, et~al.
\newblock {OpenScene}: {3D} scene understanding with open vocabularies.
\newblock In {\em IEEE/CVF Conference on Computer Vision and Pattern Recognition (CVPR)}, pages 815--824, 2023.

\bibitem{qin2024langsplat}
Minghan Qin, Wanhua Li, Jiawei Zhou, Haoqian Wang, and Hanspeter Pfister.
\newblock {LangSplat}: {3D} language {Gaussian} splatting.
\newblock In {\em IEEE/CVF Conference on Computer Vision and Pattern Recognition (CVPR)}, pages 20051--20060, 2024.

\bibitem{radford2021learning}
Alec Radford, Jong~Wook Kim, Chris Hallacy, Aditya Ramesh, Gabriel Goh, Sandhini Agarwal, Girish Sastry, Amanda Askell, Pamela Mishkin, Jack Clark, et~al.
\newblock Learning transferable visual models from natural language supervision.
\newblock In {\em International Conference on Machine Learning (ICLM)}, pages 8748--8763. PMLR, 2021.

\bibitem{raman2022planning}
Shreyas~Sundara Raman, Vanya Cohen, Eric Rosen, Ifrah Idrees, David Paulius, and Stefanie Tellex.
\newblock Planning with large language models via corrective re-prompting.
\newblock In {\em NeurIPS Foundation Models for Decision Making Workshop (FMDM)}, 2022.

\bibitem{ranasinghe2022perceptual}
Kanchana Ranasinghe, Brandon McKinzie, Sachin Ravi, Yinfei Yang, Alexander Toshev, and Jonathon Shlens.
\newblock Perceptual grouping in contrastive vision-language models.
\newblock In {\em IEEE/CVF International Conference on Computer Vision (ICCV)}, pages 5548--5561, 2023.

\bibitem{reiser2021kilonerf}
Christian Reiser, Songyou Peng, Yiyi Liao, and Andreas Geiger.
\newblock {KiloNeRF}: Speeding up neural radiance fields with thousands of tiny {MLPs}.
\newblock In {\em IEEE/CVF International Conference on Computer Vision (ICCV)}, pages 14335--14345, 2021.

\bibitem{song2023llm}
Chan~Hee Song, Jiaman Wu, Clayton Washington, Brian~M Sadler, Wei-Lun Chao, and Yu Su.
\newblock {LLM-Planner}: Few-shot grounded planning for embodied agents with large language models.
\newblock In {\em IEEE/CVF International Conference on Computer Vision (ICCV)}, pages 2998--3009, 2023.

\bibitem{sucar2021imap}
Edgar Sucar, Shikun Liu, Joseph Ortiz, and Andrew~J Davison.
\newblock {iMAP}: Implicit mapping and positioning in real-time.
\newblock In {\em IEEE/CVF International Conference on Computer Vision (ICCV)}, pages 6229--6238, 2021.

\bibitem{takikawa2021neural}
Towaki Takikawa, Joey Litalien, Kangxue Yin, Karsten Kreis, Charles Loop, Derek Nowrouzezahrai, Alec Jacobson, Morgan McGuire, and Sanja Fidler.
\newblock Neural geometric level of detail: Real-time rendering with implicit {3D} shapes.
\newblock In {\em IEEE/CVF Conference on Computer Vision and Pattern Recognition (CVPR)}, pages 11358--11367, 2021.

\bibitem{wang2022go}
Jingwen Wang, Tymoteusz Bleja, and Lourdes Agapito.
\newblock {GO-Surf}: Neural feature grid optimization for fast, high-fidelity {RGB-D} surface reconstruction.
\newblock In {\em International Conference on 3D Vision (3DV)}, pages 433--442. IEEE, 2022.

\bibitem{yang2023sam3d}
Yunhan Yang, Xiaoyang Wu, Tong He, Hengshuang Zhao, and Xihui Liu.
\newblock {SAM3D}: Segment anything in {3D} scenes.
\newblock {\em arXiv preprint arXiv:2306.03908}, 2023.

\bibitem{zhi2021ilabel}
Shuaifeng Zhi, Edgar Sucar, Andre Mouton, Iain Haughton, Tristan Laidlow, and Andrew~J Davison.
\newblock {iLabel}: Interactive neural scene labelling.
\newblock {\em arXiv preprint arXiv:2111.14637}, 2021.

\bibitem{zhong2023shine}
Xingguang Zhong, Yue Pan, Jens Behley, and Cyrill Stachniss.
\newblock {SHINE-Mapping}: Large-scale {3D} mapping using sparse hierarchical implicit neural representations.
\newblock In {\em IEEE International Conference on Robotics and Automation (ICRA)}, pages 8371--8377, 2023.

\bibitem{zhu2024loopsplat}
Liyuan Zhu, Yue Li, Erik Sandstr{\"o}m, Konrad Schindler, and Iro Armeni.
\newblock {LoopSplat}: Loop closure by registering {3D} {Gaussian} splats.
\newblock {\em arXiv preprint arXiv:2408.10154}, 2024.

\bibitem{zhu2024semgauss}
Siting Zhu, Renjie Qin, Guangming Wang, Jiuming Liu, and Hesheng Wang.
\newblock {SemGauss-SLAM}: Dense semantic {Gaussian} splatting slam.
\newblock {\em arXiv preprint arXiv:2403.07494}, 2024.

\bibitem{zhu2022nice}
Zihan Zhu, Songyou Peng, Viktor Larsson, Weiwei Xu, Hujun Bao, Zhaopeng Cui, Martin~R Oswald, and Marc Pollefeys.
\newblock {NICE-SLAM}: Neural implicit scalable encoding for {SLAM}.
\newblock In {\em IEEE/CVF Conference on Computer Vision and Pattern Recognition (CVPR)}, pages 12786--12796, 2022.

\end{thebibliography}
}

\end{document}